\documentclass[letterpaper, 10 pt, conference]{ieeeconf}  % Comment this line out if you need a4paper

\IEEEoverridecommandlockouts                              % This command is only needed if 
\usepackage{graphics} % for pdf, bitmapped graphics files
\usepackage{epsfig} % for postscript graphics files
\usepackage{mathptmx} % assumes new font selection scheme installed
\usepackage{times} % assumes new font selection scheme installed
\usepackage{amsmath} % assumes amsmath package installed
\usepackage{amssymb}  % assumes amsmath package installed
\usepackage{xcolor}
\usepackage{caption}
\usepackage{subcaption} %For sub-figures
\usepackage{gensymb}
\usepackage[noadjust]{cite}

\renewenvironment{align*}{\align}{\endalign}

\title{\LARGE \bf
Magnetic-Visual Sensor Fusion-based Dense 3D Reconstruction and 
Localization for Endoscopic Capsule Robots
}

\author{Mehmet Turan$^{1}$, Yasin Almalioglu$^{2}$, Evin Pinar Ornek$^{3}$, Helder Araujo$^{4}$, Mehmet Fatih Yanik $^{5}$, and Metin Sitti$^{6}$% <-this % stops a space
\thanks{$^{1}$Mehmet Turan is with the Physical Intelligence Department, Max Planck Institute for Intelligent Systems, Germany
        {\tt\small turan@is.mpg.de}}%
\thanks{$^{2}$Yasin Almalioglu is with the Computer Science Department, University of Oxford, Oxford, UK
        {\tt\small yasin.almalioglu@cs.ox.ac.uk}}%
\thanks{$^{3}$Evin Pinar Ornek is with the Informatics Department, Technical University of Muenich, Germany
        {\tt\small evin.oernek@tum.de}}% 
\thanks{$^{4}$Helder Araujo is with the Institute for Systems and Robotics, University of Coimbra, Portugal
        {\tt\small helder@isr.uc.pt}}%
\thanks{$^{5}$M. Fatih Yanik is with the Department of Information Technology and Electrical Engineering, Zurich, Switzerland
        {\tt\small yanik@ethz.ch}}%
\thanks{$^{6}$Metin Sitti is with the Physical Intelligence Department, Max Planck Institute for Intelligent Systems, Germany
        {\tt\small sitti@is.mpg.de}}%
}

\begin{document}

\maketitle
\thispagestyle{empty}
\pagestyle{empty}

%%%%%%%%%%%%%%%%%%%%%%%%%%%%%%%%%%%%%%%%%%%%%%%%%%%%%%%%%%%%%%%%%%%%%%%%%%%%%%%%
\begin{abstract}
Reliable and real-time 3D reconstruction and localization functionality is a crucial prerequisite for the navigation of actively controlled capsule endoscopic robots as an emerging, minimally invasive diagnostic and therapeutic technology for use in the gastrointestinal (GI) tract. In this study, we propose a fully dense, non-rigidly deformable, strictly real-time, intraoperative map fusion approach for actively controlled endoscopic capsule robot applications which combines magnetic and vision-based localization, with non-rigid deformations based frame-to-model map fusion. The performance of the proposed method is demonstrated using four different ex-vivo porcine stomach models. Across different trajectories of varying speed and complexity, and four different endoscopic cameras, the root mean square surface reconstruction errors $1.58$ to $2.17$ cm.

\end{abstract}

\section{Introduction}

%%%%%%%%%%%%%%%%%%%%%%%%%%%%%%%%%%%%%%%%%%%%%%%%%%%%%%%%%%%%%%%%%%%%%%%%%%%%%%%%

Gastrointestinal diseases are the primary diagnosis for about 28 million patient visits per year in the United States\cite{NCHS}.
In many cases, endoscopy is an effective diagnostic and therapeutic tool, and as a result about 7 million upper and 11.5 million lower endoscopies are carried out each year in the U.S.\ \cite{peery2012burden}.  
Wireless capsule endoscopy (WCE), introduced in 2000 by Given Imaging Ltd., has revolutionized patient care by enabling inspection of regions of the GI tract that are inaccessible with traditional endoscopes, and also by reducing the pain associated with traditional endoscopy \cite{iddan2000wireless}. 
Going beyond passive inspection, researchers are striving to create capsules that perform active locomotion and intervention \cite{moglia2007wireless}.
With the integration of further functionalities, e.g. remote control, biopsy, and embedded therapeutic modules, WCE can become a key technology for GI diagnosis and treatment in near future. 

\begin{figure}
\centering
  	\includegraphics[width=\columnwidth]{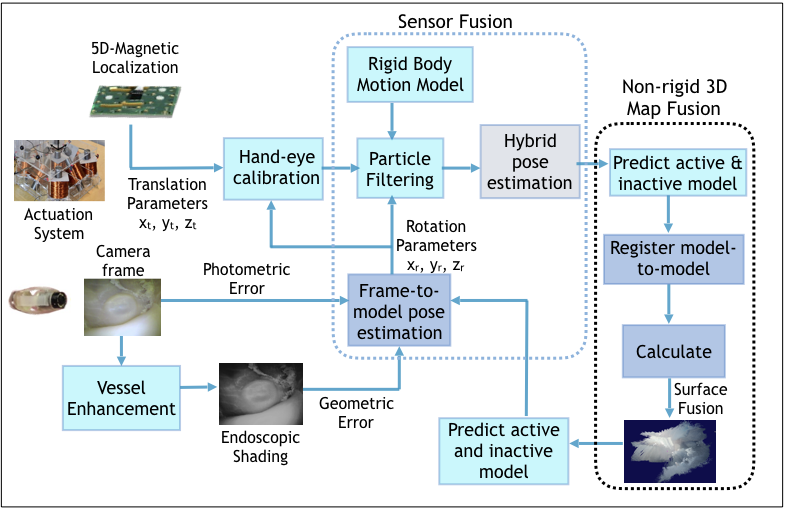}
	\caption{System overview including 5-DoF magnetic localization, 6-DoF visual joint photometric-geometric frame-to-model pose optimization, inter-sensor calibration, particle filtering based sensor fusion, non-rigid deformations based frame-to-model map fusion.}
\label{fig:model_flowchart}     
\end{figure}

Several research groups have recently proposed active, remotely controllable robotic capsule endoscope prototypes equipped with additional operational functionalities, such as highly localized drug delivery, biopsy, and other medical functions \cite{sitti2015biomedical, turanendovmfuse2017, TURAN20181861, turanendosensor2017, Turan2017, Turan2018, turan2017endovo, DBLP:journals/corr/TuranAKS17, DBLP:journals/corr/TuranPJAKS17, DBLP:journals/corr/TuranAJAKS17, DBLP:journals/corr/TuranAAKS17}. 
To facilitate effective navigation and intervention, the robot must be accurately localized and must also accurately perceive the surrouding tissues. 
Three-dimensional intraoperative SLAM algorithms will therefore be an indispensable component of future active capsule systems. Several localization methods have been proposed for robotic capsule endoscopes such as fluoroscopy \cite{than2012review}, ultrasonic imaging \cite{yim20133}, positron emission tomography (PET) \cite{than2012review}, magnetic resonance imaging (MRI) \cite{than2012review}, radio transmitter based techniques, and magnetic field-based techniques \cite{ son20165}. 
It has been proposed that combinations of sensors, such as {RF} range estimation and visual odometry, may improve the estimation accuracy \cite{geng2015accuracy}. Morover, solutions that incorporate vision are attractive because a camera is already present on capsule endoscopes, and vision algorithms have been widely applied for robotic localization and map reconstruction. 
 
\begin{figure*}
	\includegraphics[width=\textwidth]{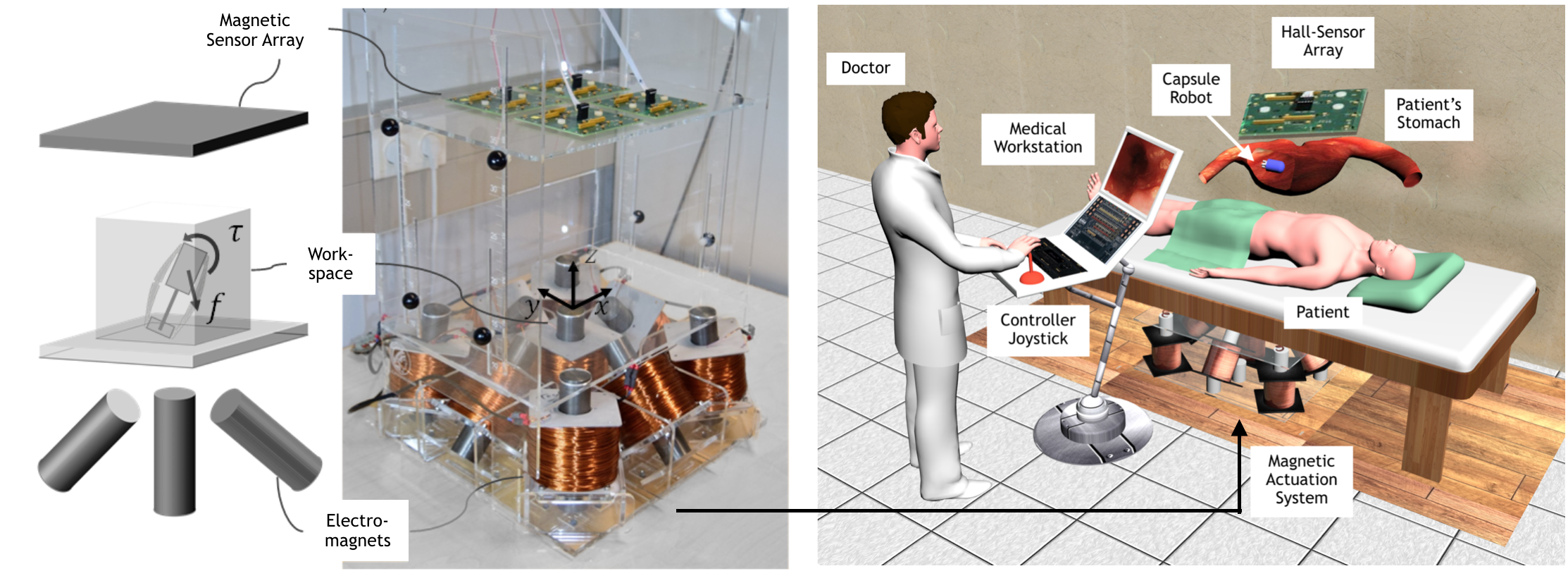}
	\caption{Demonstration of the active endoscopic capsule robot operation using MASCE (Magnetically actuated soft capsule endoscope) designed for disease detection, drug delivery and biopsy-like operations in the upper GI-tract. MASCE is composed of a RGB camera, a permanent magnet, an empty space for drug chamber and a biopsy tool. Electromagnetic coils based actuation unit below the patient table excerts forces and torques to execute the desired motion. Medician operates the screening, drug delivery and biosy processes in real-time using the live video stream onto the medical workstation and the controller joystick to manevour the endoscopic capsule to the desired position/orientation and to execute desired therapeutic actions such as drug release and biopsy. Actuation system of the MASCE: The magnet exerts magnetic force and torque on the capsule in response to a controlled external magnetic field. The magnetic torque and forces are used to actuate the capsule robot and to release drug. Magnetic fields from the electromagnets generate the magnetic force and torque on the magnet inside MASCE so that the robot moves inside the workspace. Sixty-four three-axis magnetic sensors are placed on the top, and nine electromagnets are placed in the bottom.}
	\label{fig:actuation_system}
\end{figure*}

Feature-based SLAM methods have been applied on endoscopic type of image sequences in the past e.g \cite{mountney2009dynamic,grasa2014visual,stoyanov2010real, liu2009capsule, turanendovmfuse2017, turanendosensor2017, Turan2017, Turan2018, turan2017endovo}. As improvements to accomodate the flexibility of the GI tract, \cite{mountney2010motion} suggested a motion compensation model to deal with peristaltic motions, whereas \cite{mountney2006simultaneous} proposed a learning algorithm to deal with them. \cite{lin2013simultaneous} adapted parallel tracking and mapping techniques to a stereo-endoscope to obtain reconstructed 3D maps that were denser when compared to monoscopic camera methods.  \cite{mahmoud2016orbslam} has applied ORB features to track the camera and proposed a method to densify the reconstructed 3D map, but pose estimation and map reconstruction are still not accurate enough.
All of these methods can fail to produce accurate results in cases of 
low texture areas, motion blur, specular highlights, and sensor noise 
-- all of which are typically present during endoscopy. 
In this paper, we propose that a non-rigidly deformable RGB Depth fusion 
method, which combines magnetic localization and visual pose 
estimation using particle filtering, can provide real-time, accurate 
localization and mapping for endoscopic capsule robots. 
We demonstrate the system in four different ex-vivo porcine stomachs by measuring its performance in terms of both surface mapping and capsule localization accuracy.

\section{System Overview and Analysis}

The system architecture of the method is depicted in Figure \ref{fig:model_flowchart}. Alternating between localization and mapping, our approach performs frame-to-model 3D map reconstruction in real-time. Below we summarize key steps of the proposed system:

\begin{itemize}

\item Estimate 3D position of the endoscopic capsule robot pose using magnetic localization system;

\item Estimate 3D rotation of the endoscopic capsule robot pose using visual joint photometric-geometric frame-to-model pose optimization;

\item Perform offline inter-sensor calibration between magnetic hall sensor array and capsule camera system; 

\item Fuse magnetic position and visual rotation information using particle filtering and 6-DoF rigid body motion model;

\item Perform non-rigid frame-to-model map registration making use of hybrid magneto-visual pose estimation and deformation constraints defined by the graph equations; 
\item In case there exists an intersection of the active model with the inactive model within the current frame, fuse intersecting regions and deform the entire model non-rigidly.

\end{itemize}

\section{Method}

\begin{figure*}[t]

	\centering
	\begin{subfigure}[t]{0.23\textwidth} 
	    \includegraphics[width=1\textwidth]{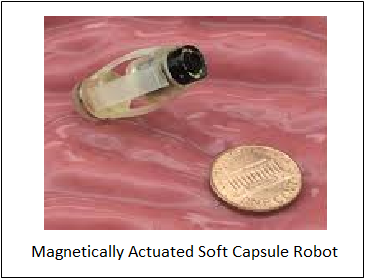}
        %\caption{Original image} \label{fig:b}
	\end{subfigure}
	~
	\begin{subfigure}[t]{0.23\textwidth} 
	    \includegraphics[width=1\textwidth]{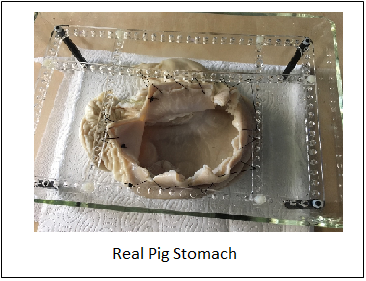}
        %\caption{Without fine-tuning (KITTI)} \label{fig:a}
	\end{subfigure}
	~
	\begin{subfigure}[t]{0.23\textwidth} 
	    \includegraphics[width=1\textwidth]{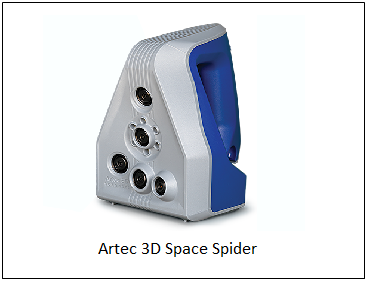}
        %\caption{Original image} \label{fig:b}
	\end{subfigure}
	~
	\begin{subfigure}[t]{0.23\textwidth} 
	    \includegraphics[width=1\textwidth]{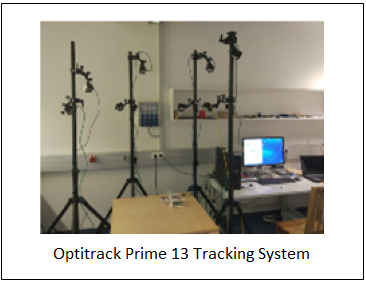}
        %\caption{After transfer learning} \label{fig:c}
	\end{subfigure}
	~
	\caption{ Illustration of the experimental setup. MASCE is a magnetically actuated robotic capsule endoscope prototype which has a ringmagnet on the body. An electromagnetic coil array consisting of nine coils is used for the actuation of the MASCE. An opened and oiled porcine stomach simulator is used to represent human stomach. Artec 3D scanner is used for ground truth map estimation. 
	OptiTrack system consisting of eight infrared cameras is employed for the ground truth pose estimation.}
    \label{fig:exp_setup}
	
\end{figure*}

\subsection{Magnetic Localization System}
Our 5-DoF magnetic localization system is designed for the position and orientation estimation of untethered mesoscale magnetic robots \cite{son20165}.
The system uses an external magnetic sensor system and electromagnets for the localization of the magnetic capsule robot. 
A 2D-Hall-effect sensor array measures the component of the magnetic field from the permanent magnet inside the capsule robot at several locations outside of the robotic workspace. 
Additionally, a computer-controlled magnetic coil array consisting of nine electromagnets generates the magnetic field for actuation. 
The core idea of our localization technique is the separation of the capsule's magnetic field component from the actuator's magnetic field component. 
For that purpose, the actuator's magnetic field is subtracted from the magnetic field data which is acquired by a Hall-effect sensor array. 
As a further step, second-order directional differentiation is 
applied to reduce the localization error. The magnetic localization 
system estimates a 5-DoF pose, which includes 3D translation and 
rotation about two axes. (From the magnetic localization information, our system only uses the 3D 
position parameters and the scale information). 

%\begin{figure}[t!]
%	\centering
%	\includegraphics[width=\columnwidth]{figures/actuation.png}
%	\caption{Actuation system of the MASCE: The magnet exerts magnetic force and torque on the capsule in response to a controlled external magnetic field. The magnetic torque and forces are used to actuate the capsule robot and to release drug. Magnetic fields from the electromagnets generate the magnetic force and torque on the magnet inside MASCE so that the robot moves inside the workspace. Sixty-four three-axis magnetic sensors are placed on the top, and nine electromagnets are placed in the bottom.}
%	\label{fig:actuation_system}
%\end{figure}

\subsection{Visual Localization}
We propose the use of a direct surfel map fusion method for actively controllable endoscopic capsule robots. 
The core algorithm is inspired by and modified from the ElasticFusion method originally described by Whelan et al. \cite{whelan2016elasticfusion}, which uses a dense map and non-rigid model deformation to account for changing environments.
It performs joint volumetric and photometric alignment, frame-to-model predictive tracking, and dense model-to-model loop closure with non-rigid space deformation. 
Prior to using endoscopic video with such a method, the images must first be prepared.
\subsubsection{Multi-scale vessel enhancement and depth image creation}
\begin{figure}
	\centering
	\includegraphics[width=1\linewidth]{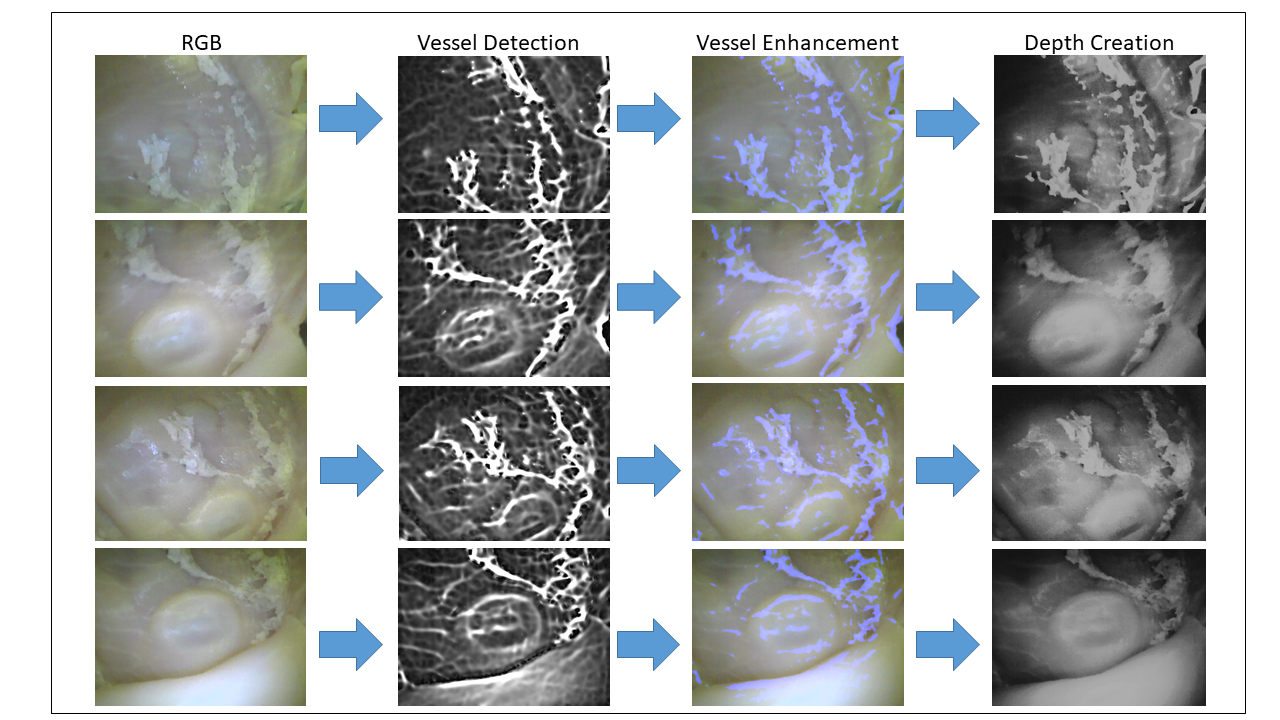}
	\caption{ For a given RGB frame, we extract the Hessian matrix and derive its eigenvalues and principal directions to detect the vessel. We convolve the input frame and final output to create multiple scale representations to identify the different vessels. After enhancement of vessel detected frame, we use shape from shading to create depth map. Qualitative results for sample frames are illustrated in the figure. Here, the dataset of our samples are collected in our experimental setup from an ex-vivo real pig stomach. }
	\label{fig:datasam}
\end{figure}

Endoscopic images have mostly homogeneous and poorly textured areas. To prepare the camera frames for input into the ElasticFusion pipeline, our framework starts with a vessel enhancement operation inspired from \cite{frangi1998multiscale}. 
Our approach enhances blood vessels by analyzing the multiscale second order local structure of an image. First, we extract the Hessian matrix :
\begin{equation}
	H=\begin{bmatrix}
			I_{xx}& I_{xy}\\ 
			I_{yx}& I_{yy}
		\end{bmatrix}
		\end{equation}
where $I$ is the input image, and $I_{xx}$, $I_{xy}$, $I_{yx}$, $I_{yy}$ the second order derivatives, respectively. Secondly, eigenvalues $\left | \lambda_{1} \right |\leq  \left | \lambda_{2} \right |$ and principal directions $u_{1}$,  $u_{2}$ of the Hessian matrix are extracted. The eigenvalues and principal directions are then ordered and analyzed to decide whether the region belongs to a vessel. To identify vessels in different scales and sizes, multiple scales are created by convolving the input image and the final output is taken as the maximum of the vessel filtered image across all scales. Figure \ref{fig:datasam} shows input RGB images, vessel detection and vessel enhancement results for four different frames.
 
To create depth from input RGB data, we implemented a real-time 
version of the perspective shape from shading under realistic 
conditions \cite{pers} by reformulating the complex inverse problem into a highly parallelized non-linear optimization problem, which we solve efficiently using GPU programming and a Gauss-Newton solver. Figure \ref{fig:datasam} shows samples of input RGB images and depth images created from them.

\subsubsection{Joint photometric-geometric pose estimation}

The vision-based localization system operates on the principle of optimizing both relative photometric and geometric pose errors between consecutive frames. The camera pose of the endoscopic capsule robot is described by a transformation matrix $\textbf{P}_t$:

\begin{equation}
\textbf{P}_t = \begin{bmatrix}
       \textbf{R}_t & \textbf{t}_t            \\[0.3em]
       0_{1\times 3} & 1           
     \end{bmatrix}
     \in \mathbb{SE}_3.
\end{equation}

Given the depth image $\mathcal{D}$, the 3D back-projection of a point $\textbf{u}$ is defined as $\textbf{p}(\textbf{u},\mathcal{D}) 
= \textbf{K}^{-1}\textbf{\textit{u}}d(\textbf{u})$, where  
$\textbf{K}$ is the camera intrinsics matrix and 
$\textbf{\textit{u}}$ is the homogeneous form of $\textbf{u}$.
Geometric pose estimation is performed by minimizing the energy cost function $E_{icp}$ between the current depth frame, $\mathcal{D}^l_t$, and the active depth model, $\hat{\mathcal{D}}_{t-1}^a$:
\begin{equation}
E_{icp} = \sum_k{((\textbf{v}^k - \exp{(\hat{\xi})}\textbf{T} \textbf{v}_k^t)\cdot\textbf{n}^k)^2}
\end{equation}
where $\textbf{v}^k_t$ is the back-projection of the $k$-th vertex in $\mathcal{D}^l_t$, $\textbf{v}^k$ and $\textbf{n}^k$ are the corresponding vertex and normal from the previous frame. $\textbf{T}$ is the estimated transformation from the previous to the current robot pose and $\exp{(\hat{\xi})}$ is the exponential mapping function from Lie algebra $\mathfrak{se}_3$ to Lie group $\mathbb{SE}_3$, which represents small changes The photometric pose $\xi$ between the current surfel-based reconstructed RGB image $\mathcal{C}^l_t$ and the active RGB model $\hat{\mathcal{C}}^a_{t-1}$ is determined by minimizing the photometric energy cost function: 
\begin{equation}
E_{rgb} = \sum_{\textbf{u} \in \Omega} \left( I(\textbf{u},\mathcal{C}^l_t) - I(\pi(\textbf{K} \exp(\hat{\xi}) \textbf{T} \textbf{p}(\textbf{u} , \mathcal{D}^l_t)), \hat{\mathcal{C}}_{t-1}^a) \right)^2
\end{equation}
where as above $\textbf{T}$ is the estimated transformation from previous to the current camera pose. 

The joint photometric-geometric pose optimization is defined by the cost function:
\begin{equation}
E_\textrm{track} = E_\textrm{icp} + w_\textrm{rgb}E_\textrm{rgb},
\end{equation}
with $w_\textrm{rgb} = 0.13$, which was determined experimentally for our datasets. For the minimization of this cost function in real-time, the Gauss-Newton method is employed. At each iteration of the method, the transformation $\textbf{T}$ is updated as $\textbf{T} \to \exp(\hat{\xi})\textbf{T}$. For scene reconstruction, we use surfels. Each surfel has a position, normal, color, weight, radius, initialization timestamp and last updated timestamp. We also define a deformation graph consisting of a set of nodes and edges to detect non-rigid deformations throughout the frame sequence. Each node $\mathcal{G}^n$ has a timestamp $\mathcal{G}^n_{t_0}$, a position $\mathcal{G}_g^n \in \mathbb{R}^3$ and a set of neighboring nodes $\mathcal{N}(\mathcal{G}^n$). The directed edges of the graph are neighbors of each node. A graph is connected up to a neighbor count $k$ such that $\forall n,|\mathcal{N}(\mathcal{G}^n)| = k$. Each node also stores an affine transformation in the form of a $3\times 3$ matrix $\mathcal{G}^n_{\textbf{R}}$  and a $3\times 1$ vector $\mathcal{G}_{\textbf{t}}^n$. When deforming a surface, the $\mathcal{G}^n_{\textbf{R}}$ and $\mathcal{G}_{\textbf{t}}^n$  parameters of each node are optimized according to surface constraints. In order to apply a deformation graph to the surface, each surfel $\mathcal{M}^s$ identifies a set of influencing nodes in the graph $\mathcal{I}(\mathcal{M}^s, \mathcal{G})$. The deformed position of a surfel is given by:
\begin{equation}
	\hat{\mathcal{M}}^s_{\textbf{p}} = \phi(\mathcal{M}^s) = \sum_{n 
	\in \mathcal{I}(\mathcal{M}^s, \mathcal{G})} w^n(\mathcal{M}^s) 
	[\mathcal{G}^n_{\textbf{R}}(\mathcal{M}_{\textbf{p}}^s - 
	\mathcal{G}_{\textbf{g}}^n) + \mathcal{G}_{\textbf{g}}^n + 
	\mathcal{G}_{\textbf{t}}^n]
\end{equation}
while the deformed normal of a surfel is given by:
\begin{equation}
	\hat{\mathcal{M}}^s_{\textbf{p}} = \sum_{n \in 
	\mathcal{I}(\mathcal{M}^s, \mathcal{G})} w^n 
	(\mathcal{M}^s)\mathcal{G}^{{n-1}^T}_{\textbf{R}} 
	\mathcal{M}^s_{\textbf{n}},
\end{equation}
where $w^n (\mathcal{M}^s)$ is a scalar representing the influence of 
$\mathcal{G}^n$ on surfel $\mathcal{M}^s$, summing to a total of $1$ 
when $n = k$:
\begin{equation}
	w^n (\mathcal{M}^s) = (1 - 
	||\mathcal{M}^s_{\textbf{p}}-\mathcal{G}_{\textbf{g}}^n ||_2 / 
	d_\textrm{max})^2.
\end{equation}
Here, $d_\textrm{max}$ is the Euclidean distance to the $k+1$-nearest 
node of $M^s$. 

To ensure a globally consistent surface reconstruction, the framework closes loops with the existing map as those areas are revisited. This loop closure is performed by fusing reactivated parts of the inactive model into the active model and simultaneously deactivating surfels which have not appeared for a period of time.

\subsection{Particle Filtering based Magneto-Visual Sensor Fusion}
\label{sec:fusion}

We developed a particle filtering based sensor fusion method for endoscopic capsule robots which provides robustness against sensor failure through the introduction of latent variables characterizing the sensor's reliability as either normal or failing, which are estimated along with the system state. The method is inspired by and modifed from \cite{caron2007particle}. As motion model, we use a rigid motion model (3D rotation and 3D translation) assuming constant velocity which is fairly obeyed during incremental motions of magnetically actuated endoscopic capsule robots. The proposed fusion approach estimates the 3D translation using the 
measurements from the magnetic sensor, which include the 
scale factor, and the 3D rotation using visual information provided by the monocular endoscopic capsule camera. 

The state $\mathbf{x_{t}}$ composes the 6-DoF pose for the capsule robot, which is assumed to propagate in time according to a transition model:
\begin{equation}
\label{eq:state_transition}
\mathbf{x}_{t} = f(\mathbf{x}_{t-1} , \mathbf{v}_{t} )  
\end{equation}
where $f$ is a non-linear state transition function and  $\mathbf{v_{t}}$ is white noise. $t$ is the index of a time sequence, $t \in \{1,2,3, ...\}$. Observations of the pose are produced by $n$ sensors $\mathbf{z}_{k,t}  (k = 1,...,n)$ in general,  where the probability distribution $p(\mathbf{z}_{k,t}|\mathbf{x}_t)$ is known for each sensor.
%\begin{figure}
%	\centering
%	\includegraphics[width=0.9\linewidth]{figures/model.png}
%	\caption{The switching state-space model. The double rectangles denote observable variables and the gray rectangles denote hyper-parameters.}
%	\label{fig:graph_model}
%\end{figure}
We estimate the 6-DoF pose states relying on latent (hidden) variables by using the Bayesian filtering approach. The hidden variables of sensor states are denoted as $s_{k,t}$, which we call switch variables, where $s_{k,t}\in \{0,..., d_{k}\}$ for $k = 1,...,n$. $d_{k}$ is the number of possible observation models, e.g., failure and nominal sensor states. 
The observation model for $\mathbf{z}_{k,t}$ can be described as:

\begin{equation}
\mathbf{z}_{k,t} = h_{k, s_{k, t},t}(\mathbf{x}_{t}) + \mathbf{w}_{k, s_{k, t},t} 
\end{equation}
\begin{figure*}
\centering
	\begin{subfigure}{\columnwidth} 
		\includegraphics[width=\textwidth]{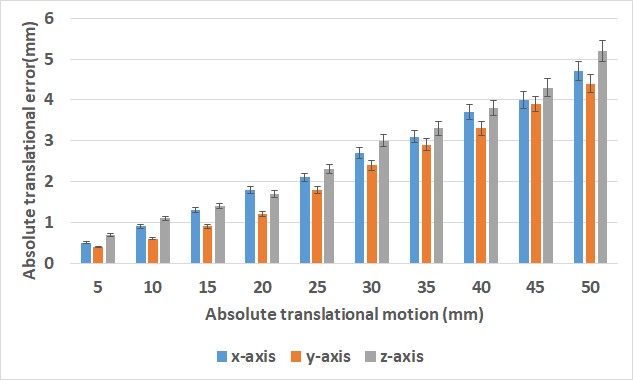}
		\caption{Translational error}
		\label{fig:err_trans}       % Give a unique label
	\end{subfigure}
	\begin{subfigure}{\columnwidth} 
		\includegraphics[width=\textwidth]{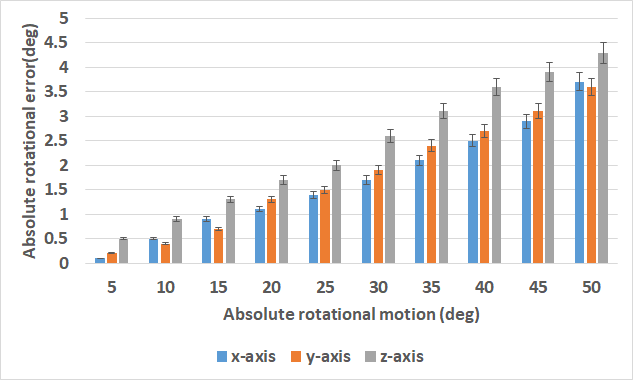}
		\caption{Rotational error}
		\label{fig:err_rot} 
	\end{subfigure}
	\begin{subfigure}{\columnwidth} 
		\includegraphics[width=\textwidth]{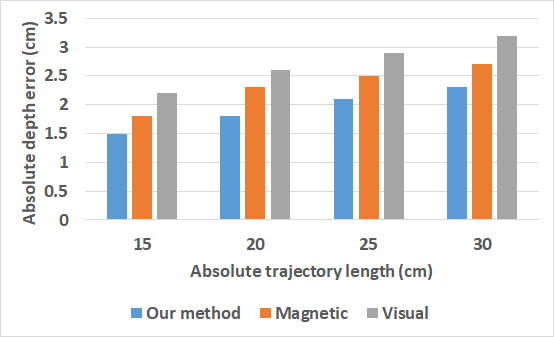}
		\caption{Depth error}
		\label{fig:err_depth} 
	\end{subfigure}
	\begin{subfigure}{\columnwidth} 
		\includegraphics[width=\textwidth]{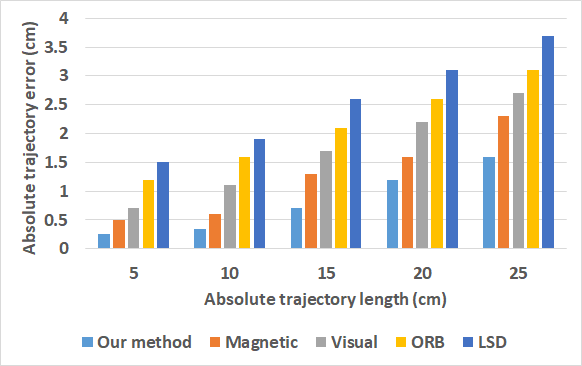}
		\caption{Trajectory error}
		\label{fig:err_traj_comp} 
	\end{subfigure}
	\caption{Figure (a) and Figure (b) demonstrates translational and rotational errors of x, y, z axes for our proposed method. The translational motion of 5 mm results in around 0.5 mm drift on average for x,y,z, whereas a 5 degree rotational motion results in 0.5 degree error maximum. The absolute depth error results for magnetic localization, visual localization and our method is illustrated in (c). It can be observed that our method outperforms the others in depth estimation for different trajectory lengths. In (d), we compare the trajectory errors of magnetic localization, visual localization, ORB SLAM, LSD SLAM and our method. For each of different trajectory lengths, our method outperforms the localization methods that use only visual or magnetic sensors and SLAM methods. For example, in a trajectory with 20 cm, our method estimates with a 1.25 cm error, whereas the error of magnetic localization is 1.6, visual localization is 2.1, ORB SLAM is 2.6, and LSD SLAM is 3. }
	\label{fig:rot_trans_error}
\end{figure*}
where $h_{k, s_{k, t},t}(\mathbf{x}_{t})$ is the non-linear observation function and $\mathbf{w}_{k, s_{k, t},t} $ is the observation noise. The latent variable of the switch parameter $s_{k, t}$ is defined to be $0$ if the sensor is in a failure state, which means that observation $\mathbf{z}_{k,t}$ is statistically independent of $\mathbf{x}_{t}$, and $1$ if the sensor $k$ is in its nominal state of work. The prior probability for the switch parameter $s_{k,t}$ being in a given state $j$, is denoted as  $\alpha_{k,j,t}$ and it is the probability for each sensor to be in a given state:

\begin{equation}
Pr(s_{k,t}=j) = \alpha_{k,j,t} , \quad 0 \leq j \leq d_{k} 
\end{equation}
where $\alpha_{k,j,t} \geq 0$ and  $\sum_{j=0}^{d_{k}}\alpha_{k,j,t}=1$ with a Markov evolution property. The objective posterior density function $p(\mathbf{x}_{0:t}, \mathbf{s}_{1:t}, \mathbf{\alpha}_{0:t}|\mathbf{z}_{1:t})$ and the marginal posterior probability $p(\mathbf{x}_{t}|\mathbf{z}_{1:t})$ , in general, cannot be determined in a closed form due to its complex shape. However, sequential Monte Carlo methods (\textit{particle filters}) provide a numerical approximation of the posterior density function with a set of samples (\textit{particles}) weighted by the kinematics and observation models.

\subsubsection*{Sensor Failure Detection and Handling}

\begin{figure*}
\centering
\begin{subfigure}[t]{.47\textwidth}
\includegraphics[width=\textwidth]{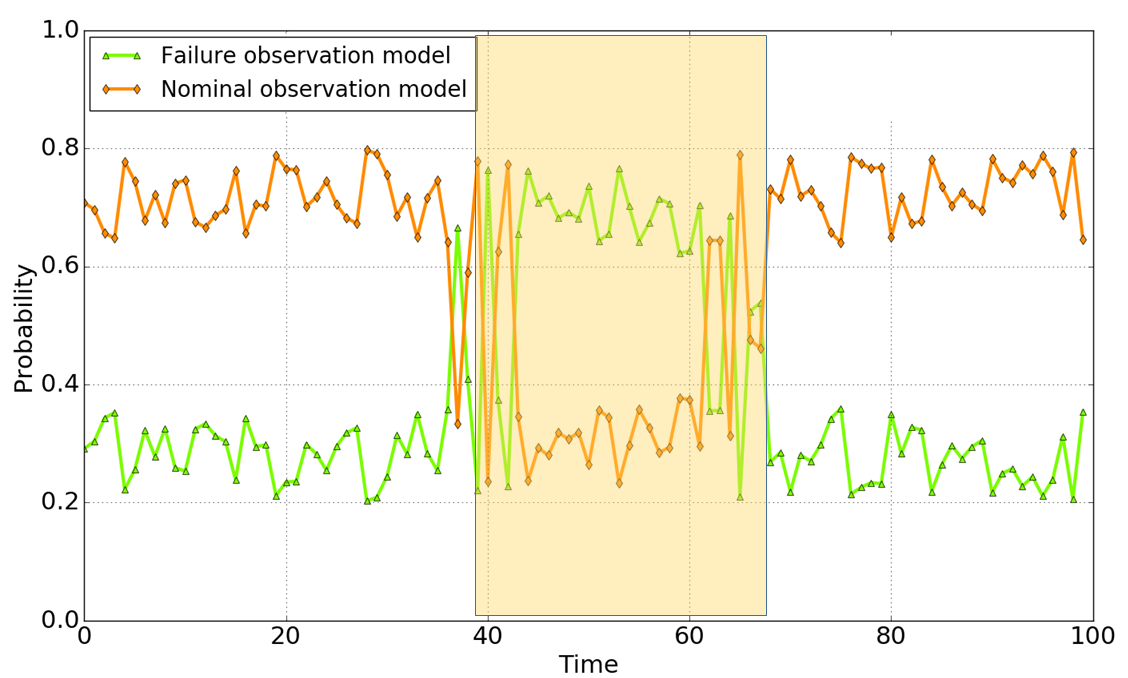}
%\caption{Translation loss decreases as the training proceeds.}
\label{fig:alpha1_result}       % Give a unique label
\end{subfigure} %
~ 
\begin{subfigure}[t]{0.47\textwidth}
\includegraphics[width=\textwidth]{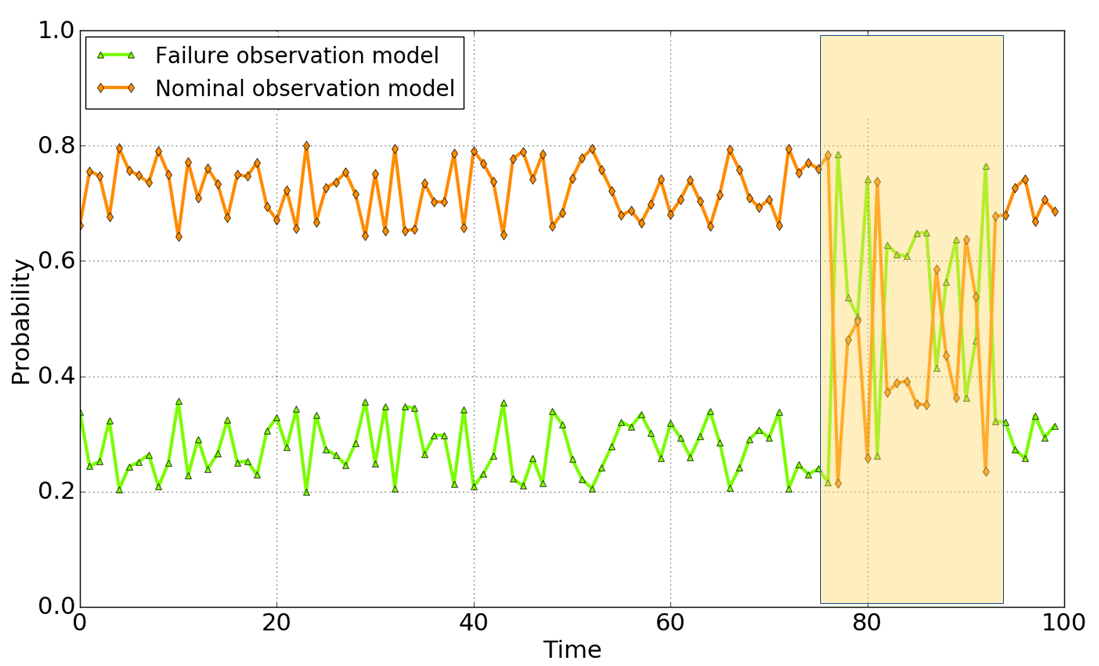}
%\caption{As the training proceeds, the rotation error decreases.}
\label{fig:alpha2_result}       % Give a unique label
\end{subfigure}

\caption{The minimum mean square error (MMSE) of $\mathbf{\alpha}_{k,t}$ for endoscopic RGB camera (left) and for magnetic localization system (right).  The switch parameter, $\mathbf{s}_{k,t}$, and the confidence parameter $\mathbf{\alpha}_{k,t}$ reflect the failure times accurately: Visual localization fails between $39-68$ seconds and magnetic localization fails between $78-92$ seconds. Both failures are detected confidentially.}
\label{fig:alpha_result}
\end{figure*}

The proposed multi-sensor fusion approach is able to detect the sensor failure periods and to handle the failures, accordingly. As seen in Fig. \ref{fig:alpha_result}, the posterior probabilities of the switch parameters $\mathbf{s}_{k, t}$ and the minimum mean square error (MMSE) estimates of $\mathbf{\alpha}_{k,t}$ indicate an accurate detection of sensor failure states. Visual localization failed between seconds 14-36 due to very fast frame-to-frame motions  and magnetic sensor failed between seconds 57-76 due to increased distance of the ringmagnet to the sensor array. Once a sensor failure is detected, the approach stops to use this sensor information until the failure state ends and uses prior information and rigid body motion model to predict the misssing information. Thanks to this switching option ability, MMSE is kept low during sensor failure as seen in Figure \ref{fig:alpha_result}. In our sensor failure model, we do not make a Markovian assumption for the switch variable $\mathbf{s}_{k,t}$ but we do for its prior $\mathbf{\alpha}_{k,t}$, resulting in a priori dependent on the past trajectory sections, which is more likely for the incremental endoscopic capsule robot motions. The model thus introduces a memory over the past sensor states rather than simply considering the last state. The length of the memory is tuned by the hyper-parameters $\sigma^{\alpha}_{k,t}$,  leading to a long memory for large values and vice-versa. This is of particular interest when considering sensor failures. Our system detects automatically failure states. Hence, the confidence in the vision sensor decreases when visual localization fails recently due to occlusions, fast-frame-to frame changes etc. On the other hand, the confidence in magnetic sensor decreases if the magnetic localization fails due to noise interferences from environment and/or if the ringmagnet has a big distance to the magnetic sensor array.

\subsection{Relative pose of magnetic and visual localization systems}
\label{sec:calib}

To relate the magnetic actuation and localization system (which is seen in Fig. \ref{fig:actuation_system}) with the proposed vision  system, the relative pose has to be estimated. The relative pose can be estimated using rigid motion from the capsule and the constraint of the rigid transformation between the magnetic sensor coordinate system and the camera coordinate system (as in eye-in-hand calibration). The vision system measures the pose of the camera, and the magnetic localization system measures the 5D pose of the magnet on the MASCE. The transformation between the coordinate frames attached to the ringmagnet and to the camera origin must be known, because the particle filter assumes that the two systems make measurements on the same system state, which in this case is a single rigid body pose associated with the capsule. In this case the magnetic system provides a 5-DoF pose while the vision system yields a 6-DoF pose. To estimate the relative pose we assumed a value for the missing rotational DoF in the magnetic sensor data and used an approach based on the method described in \cite{lebraly2010nonoverlap}. Several motions were performed, and using the estimates of the relative pose (between consecutive positions), the rigid transformation between the two coordinate systems was estimated. The use of several motions allowed the estimation of the uncertainty in the parameters.  

\begin{figure*}
\centering
	\begin{subfigure}{\columnwidth} 
		\includegraphics[width=\textwidth]{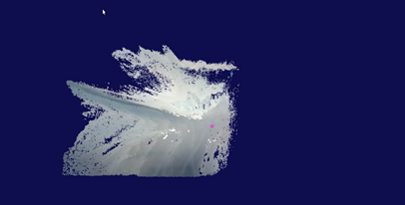}
		\caption{10 frames}
		\label{fig:surface1} 
	\end{subfigure}
	\begin{subfigure}{\columnwidth} 
		\includegraphics[width=\textwidth]{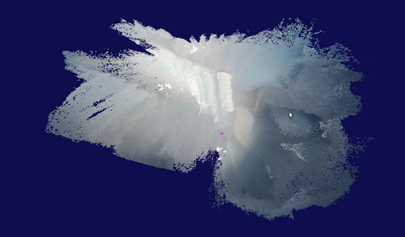}
		\caption{100 frames}
		\label{fig:surface2} 
	\end{subfigure}
	\begin{subfigure}{\columnwidth} 
		\includegraphics[width=\textwidth]{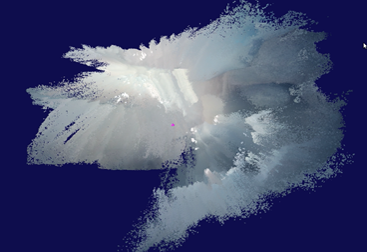}
		\caption{300 frames}
		\label{fig:surface3} 
	\end{subfigure}
	\begin{subfigure}{\columnwidth} 
		\includegraphics[width=\textwidth]{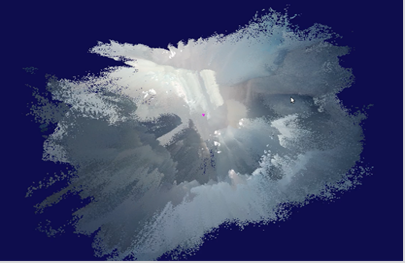}
		\caption{500 frames}
		\label{fig:surface4} 
	\end{subfigure}
	\caption{Reconstructed 3D map of a porcine non-rigid stomach simulator for total number of 10, 100, 300 and 500 frames, respectively. The illustrations are complementary to surface reconstruction errors given in Fig. \ref{fig:err_traj_comp}. It is observable that the proposed method reconstructs 3D organ surface precisely .}
	\label{fig:surface_stomach}
\end{figure*}

\section{EXPERIMENTS AND RESULTS} \label{sec:experiments}

We evaluate the performance of our system both quantitatively and qualitatively in terms of surface reconstruction, trajectory estimation and computational performance.  Figure \ref{fig:exp_setup} illustrates our experimental setup. Four different endoscopic cameras were used to capture endoscopic capsule videos which were mounted on our magnetically activated soft capsule endoscope (MASCE) systems. The dataset was recorded on four different open non-rigid porcine stomach. Ground truth 3D reconstructions of stomachs were acquired by scanning with a high-quality 3D scanner Artec Space Spider. These 3D scans served as the gold standard for the evaluations of the 3D map reconstruction. To obtain the ground truth for 6-DoF camera pose, an OptiTrack motion tracking system consisting of eight infrared cameras was utilized.  A total of 15 minutes of stomach videos were recorded containing over $10K$ frames. Some sample frames of the dataset are shown in Fig. \ref{fig:datasam} for visual reference.

\subsection{Surface reconstruction and trajectory estimation} \label{sec:trajectory_estimation}

For the duration of the pose and map reconstruction evaluations, we have only utilized sequences where the Bayesian filtering algorithm confirmed that camera and magnetic sensor remained in the nominal sensor state. We used the map benchmarking technique proposed by \cite{handa2014benchmark} for the evaluation of the map reconstruction and ATE \cite{mur2015orb} for trajectory comparisons.
Since iterative closest point algorithm (ICP) is a non-convex procedure highly dependent on a good initialization, we first manually align reference and estimated point cloud by picking six corresponding point pairs between both point clouds. Using these six manually picked corresponding point pairs, the transformation matrix is estimated which minimizes square sum difference between aligned and reference cloud. As a next step, ICP is applied between manually aligned cloud pair to fine-tune the alignment. The termination criteria for ICP iterations is an RMSE difference of 0.001 cm between consecutive iterations. We use Euclidean distances between aligned and reference cloud points to calculate the RMSE for depth. Surface reconstruction errors are compared with the magnetic localization-based and visual localization-based surface reconstruction errors in Fig. \ref{fig:err_depth}. Results indicate that the proposed method reconstructs 3D organ surface very precisely outperforming both methods. Table \ref{tab:rec_stomach} shows the reconstruction error metrics for full trajectory lengths and four different porcine stomachs including mean, median, standard deviation, minimum and maximum error. Sample 3D reconstructed maps for different lengths of frame sequences (10, 100, 300, 500 frames) are shown in Fig. \ref{fig:surface_stomach}, for visual reference. 

Figures \ref{fig:err_trans} and \ref{fig:err_rot} demonstrate absolute translational and rotational errors for our method, magnetic sensor-based localization and vision-based localization. Observation shows that proposed hybrid approach outperforms both sensor types clearly in terms of translational and rotational motion estimation.  A translational motion of 5 mm results in a drift of around 0.5 mm on average for x,y,z axes, whereas a 5 degree rotational motion results in a maximum error of 0.5 degree. Figure \ref{fig:rot_trans_error} shows the absolute trajectory errors acquired by our method, compared to ORB SLAM \cite{mur2015orb}, LSD SLAM \cite{engel2014lsd}, magnetic sensor-based and visual sensor-based localization. Results again indicate, that the proposed hybrid method outperforms other methods. For example, in a trajectory of 20 cm length, our method estimates with an error of 1.25 cm, whereas magnetic localization, visual localization, ORB and LSD SLAM estimate with an error of 1.6 cm, 2.1 cm, 2.6 cm, and 3 cm, respectively.

\begin{table}[h]
\caption{Reconstruction results for different stomach sequences.}
\label{tab:rec_stomach}
\begin{center}
\begin{tabular}{|c||c|c|c|c|}
\hline
Error (cm)& St0 & St1 & St2 & St3 \\
\hline
Mean & 1.81 & 1.97 & 1.58 & 2.17\\
\hline
Median & 1.69 & 1.55 & 1.38 & 1.98\\
\hline
Std. & 1.94 & 2.67 & 1.73 & 2.32 \\
\hline
Min & 0.00 & 0.00 & 0.00 & 0.00\\
\hline
Max & 3.4 & 4.2 & 3.1 & 4.5\\
\hline
\end{tabular}
\end{center}
\end{table}

\subsection{Computational Performance} \label{sec:computational_performance} 

To analyze the computational performance of the system, we observed the average frame processing time across the videos. The test platform was a desktop PC with an Intel Xeon E5-1660v3-CPU at 3.00 GHz, 8 cores, 32GB of RAM and an NVIDIA Quadro K1200 GPU with 4GB of memory. The execution time of the system is depended on the number of surfels in the map, with an overall average of 45 ms per frame scaling to a peak average of 52 ms implying a worst case processing frequency of 19 Hz.

\section{CONCLUSION} \label{sec:conclusion}

In this paper, we have presented a magnetic-RGB Depth fusion based 3D reconstruction and localization method for endoscopic capsule robots. Our system makes use of surfel-based dense reconstruction in combination with particle filter based fusion of magnetic and visual localization information and sensor failure detection. The proposed system is able to produce a highly accurate 3D map of the explored inner organ tissue and is able to stay close to the ground truth endoscopic capsule robot trajectory even for challenging robot trajectories. In the future, \textit{in vivo} testing is required to validate the accuracy and robustness of the approach in the challenging conditions of the GI tract. We also intend to extend our work into stereo capsule endoscopy applications to achieve even more accurate localization and mapping. In addition, an improved estimation of the relative pose between the coordinate systems of the sensors may result in improved accuracy.

%                                  % on the last page of the document manually. It shortens
%                                  % the textheight of the last page by a suitable amount.
%                                  % This command does not take effect until the next page
%                                  % so it should come on the page before the last. Make
%                                  % sure that you do not shorten the textheight too much.

%%%%%%%%%%%%%%%%%%%%%%%%%%%%%%%%%%%%%%%%%%%%%%%%%%%%%%%%%%%%%%%%%%%%%%%%%%%%%%%%

%%%%%%%%%%%%%%%%%%%%%%%%%%%%%%%%%%%%%%%%%%%%%%%%%%%%%%%%%%%%%%%%%%%%%%%%%%%%%%%

%%%%%%%%%%%%%%%%%%%%%%%%%%%%%%%%%%%%%%%%%%%%%%%%%%%%%%%%%%%%%%%%%%%%%%%%%%%%%%%
%\section*{APPENDIX}

%\section*{ACKNOWLEDGMENT}

%%%%%%%%%%%%%%%%%%%%%%%%%%%%%%%%%%%%%%%%%%%%%%%%%%%%%%%%%%%%%%%%%%%%%%%%%%%%%%%%
%\clearpage 
\bibliographystyle{IEEEtran}
\bibliography{mybibfile}
\end{document}